\documentclass[letterpaper, 10 pt, conference]{ieeeconf}  

\IEEEoverridecommandlockouts                              

\usepackage{amsmath,amsfonts,amssymb}
\usepackage{cite}
\usepackage{graphicx} 
\usepackage{siunitx} 
\usepackage{soul}
\usepackage{hyperref}

\title{\LARGE \bf
Shared Autonomy via Variable Impedance Control and Virtual Potential Field\textcolor{black}{s} for Encoding Human Demonstrations*
}
\author{Shail Jadav$^{\dagger,1,2}$, Johannes Heidersberger$^{\dagger,2}$, Christian Ott$^{3,4}$ and Dongheui Lee$^{2,4}$
\thanks{*This work was supported by funding from the DFG Priority Programme (SPP 2134) Active Self \textcolor{black}{and IIT Gandhinagar Overseas Research Fellowship.}}%
\thanks{\(\dagger\)These authors contributed to this paper equally.
}%
\thanks{$^{1}$ Shail Jadav is Indian Institute of Technology Gandhinagar, India {\tt\small shail.jadav@iitgn.ac.in}}%
\thanks{$^{2}$ Shail Jadav, Johannes Heidersberger, \& Dongheui Lee are with Autonomous Systems, Technische Universität Wien
(TU Wien), Austria, {\tt\small 
 johannes.heidersberger@tuwien.ac.at, dongheui.lee@tuwien.ac.at}}%
\thanks{$^{3}$ Christian Ott is with Automation and Control Institute, Technische Universität Wien (TU Wien), Austria   {\tt\small christian.ott@tuwien.ac.at}}%
\thanks{$^{4}$ Christian Ott and Dongheui Lee are with  the Institute of Robotics and Mechatronics, German Aerospace Center (DLR), Germany.}\\
\href{https://shailjadav.github.io/SALADS/}{\color{blue} shailjadav.github.io/SALADS/}
}
\begin{document}

\maketitle
\thispagestyle{empty}
\pagestyle{empty}

\begin{abstract}
This article introduces a framework for complex human-robot collaboration tasks, such as the co-manufacturing of furniture. For these tasks, it is essential to encode tasks from human demonstration and reproduce these skills in a compliant and safe manner. Therefore, two key components are addressed in this work: motion generation and shared autonomy. We propose a motion generator based on a time-invariant potential field, capable of encoding wrench profiles, complex and closed-loop trajectories, and additionally incorporates obstacle avoidance. Additionally, the paper addresses shared autonomy (SA) which enables synergetic collaboration between human operators and robots by dynamically allocating authority. Variable impedance control (VIC) and force control are employed, where impedance and wrench are adapted based on the human-robot autonomy factor derived from interaction forces. System passivity is ensured by an energy-tank based task passivation strategy. The framework's efficacy is validated through simulations and an experimental study employing a Franka Emika Research 3 robot.
\end{abstract}

\section{INTRODUCTION}
\begin{figure}[!hbtp]
  \centering
  \includegraphics[width=\linewidth,keepaspectratio]{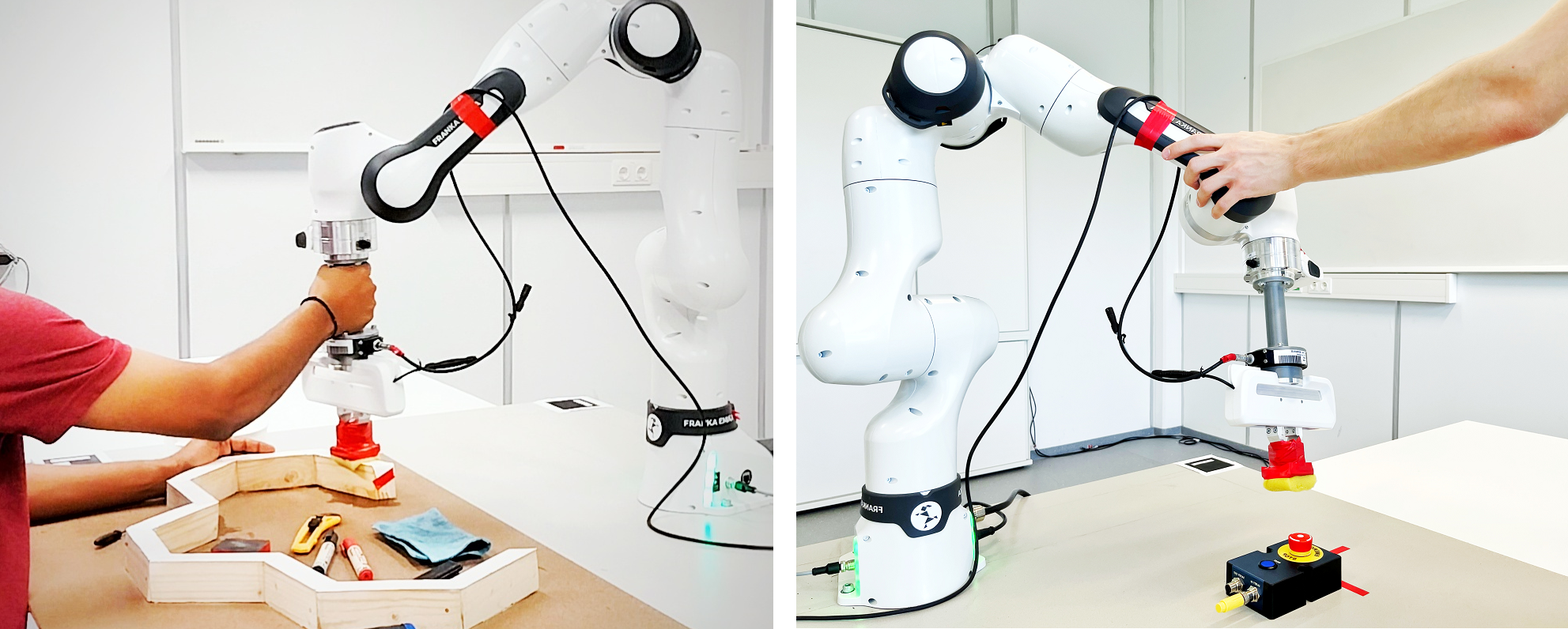}
  \vspace{-0.6cm}
  \caption{Experimental setups illustrating shared autonomy in human-robot collaboration for furniture co-manufacturing tasks (left) and button-pressing tasks (right).}
  \label{fg:setup}
  \vspace{-0.87cm}
\end{figure}

In recent years, robotic systems have transitioned from conventional applications in structured industrial environments to more dynamic human-robot collaborative scenarios, such as co-manufacturing of furniture \cite{peternel2018robot}. Such collaborative frameworks have the possibility of leveraging the advanced cognitive capabilities of humans in conjunction with the superior attributes of robotic manipulators, like high accuracy of motion repetition without fatigue, thereby enhancing overall system performance \cite{pervez2019motion,pervez2019human,balachandran2020adaptive,coelho2021whole,j2024con}. In this study, we investigate complex tasks involving human-robot collaboration (Fig. \ref{fg:setup}), such as furniture co-manufacturing, which necessitates the integration of multiple components, including motion generation algorithms informed by learning from demonstration (LfD), shared autonomy (SA) protocols to balance control between human and robot, as well as variable impedance and force control mechanisms to adapt to changing task requirements and to ensure compliant behavior. 
\par A prominent example of LfD-based motion planning for SA are dynamical systems (DS), which are known to provide stability guarantees concerning global attractors \cite{khansari2011learning}. The domain of machine learning-based DS has witnessed numerous extensions, including, but not limited to, incremental learning \cite{kronander2015incremental, Saveriano2018}, shared control \cite{khoramshahi2020dynamical}, reactivity to moving targets \cite{salehian2016dynamical}, obstacle avoidance \cite{khansari2012dynamical}. The Stable Estimator of Dynamical System (SEDS) is an algorithm based on Gaussian mixture models, offering stability guarantees by attracting towards a global point \cite{khansari2011learning}. However, there is a trade-off between stability and accuracy, often failing to precisely encode highly nonlinear demonstrations \cite{billard2022learning, Blocher2017}. Additionally, SEDS is limited to modeling trajectories that monotonically decrease in distance to the target over time \cite{nawaz2023learning}. To address this limitation, authors of \cite{pmlr-v87-figueroa18a} introduced a method based on SEDS through a Linear Parameter Varying (LPV) re-formulation of the model (LPV-DS). This approach can learn more intricate trajectories compared to the original SEDS, but it still falls short in encoding highly complex motions, especially evident in the leaf shape (see Fig. \ref{fg:comp}) within the LASA dataset of handwriting \cite{billard2022learning}. The authors in \cite{khansari2017learning} proposed an alternative approach that learns potential functions from demonstrations, addressing the limitation in precisely encoding complex shapes with the previously mentioned motion generators. While this method is capable of generating complex shapes, it is not applicable to closed-loop trajectories. Additionally, this approach employs a predefined impedance model based on the trajectory, thereby complicating the integration of dynamic authority shifts between human and robot through modulation of impedance characteristics. Furthermore, this method does not encode the requisite wrench profile for task execution, a critical component for successful human-robot collaboration that requires precise wrench modulation.

Further, SA is useful in co-manufacturing tasks with humans \cite{pichler2017towards}. There have been numerous advancements in dynamic authority allocation between human operators and robotic agents, existing methodologies often employ end-effector velocity and force, electromyography (EMG), and stability constraints for authority arbitration \cite{sharifi2021impedance}. Admittance control methodologies have been investigated in the context of shared autonomy, yet they exhibit constraints in force-production tasks involving rigid bodies, due to uncertainties induced by physical interactions \cite{fujiki2022series,landi2018passivity,kikuuwe2019torque}. 

\par Consequently, this paper presents a comprehensive framework for shared autonomy with the following contribution:
\begin{itemize}
\item We introduce a time-invariant, state-dependent motion generator that encodes task-specific trajectories and wrench profiles, accommodates complex and closed-loop paths, and integrates obstacle avoidance.
\item We propose a shared autonomy methodology that incorporates variable impedance and force control (VIC) for accurate trajectory and wrench tracking, adapting to authority levels and complying with human-induced perturbations.
\end{itemize}
Furthermore, we incorporate an energy-tank-based passivation strategy, akin to that presented in \cite{energyTankFed,schindlbeck2015unified,ENAYATI2020}, to guarantee passivity during time-varying impedance and control parameters. Finally, we validate the proposed approach through extensive experiments and simulations.

\section{Methods}
\begin{figure}[!hbtp]
  \centering
  \includegraphics[width=0.75\linewidth,keepaspectratio]{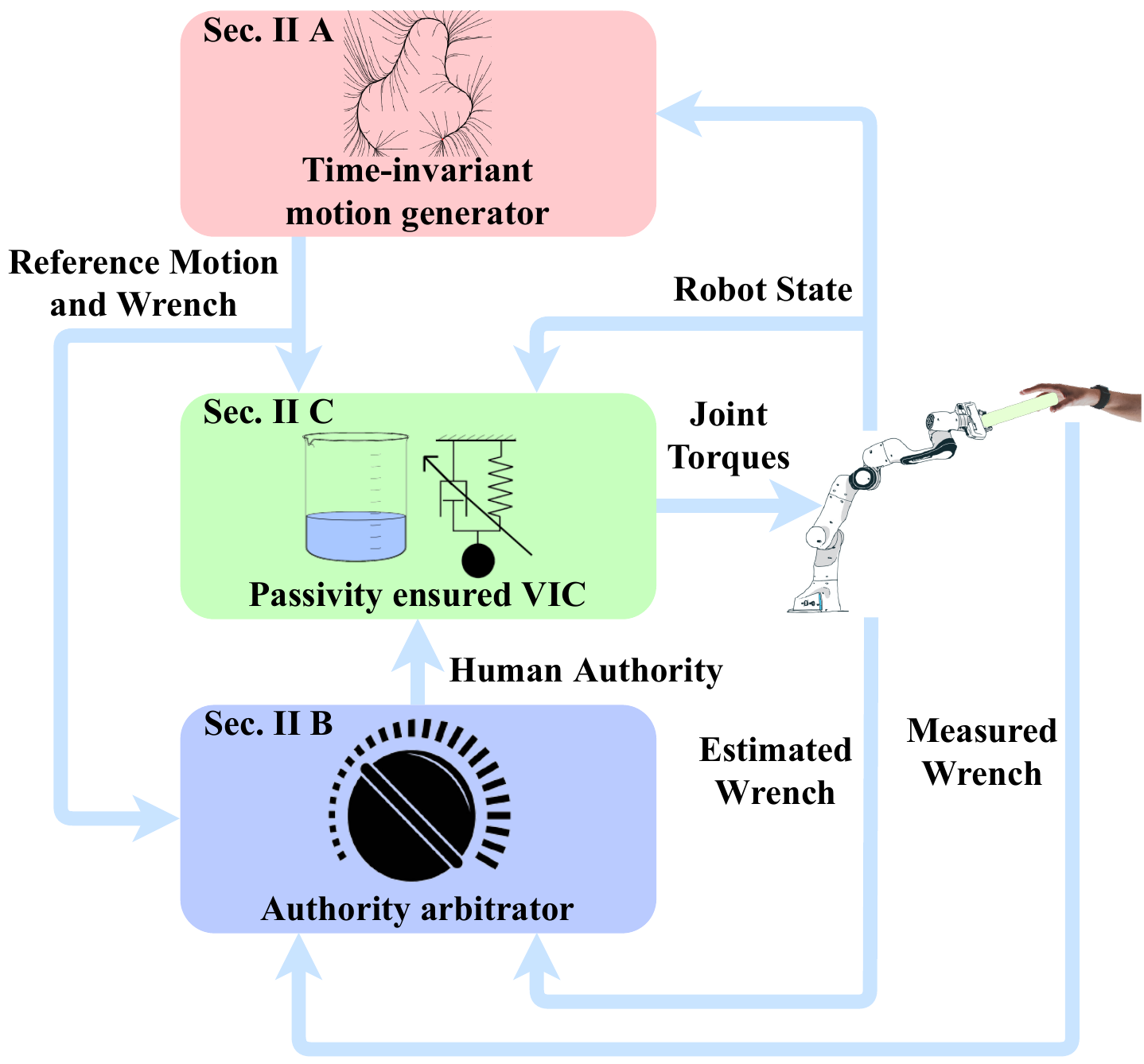}
  \vspace{-0.3cm}
  \caption{Overview of proposed framework \textcolor{black}{consisting of a motion generator, variable impedance control, and an authority arbitrator}.}
  \label{fg:bd}
  \vspace{-0.7cm}
\end{figure}

In the proposed framework, the VIC receives reference trajectories from a time-invariant, state-dependent motion generator, as shown in Fig. \ref{fg:bd}, modulated via the authority arbitrator. The authority arbitrator is modulated by human input; when force is applied by the human or any unencoded forces are detected, the compliance of the robot is increased.
\subsection{Motion Planning using Virtual Potential Fields}
Let us assume we are provided one demonstration of $\rm{T_{d}}$ datapoints consisting of the Cartesian positions of the end-effector \(\mathbf{X}_{\rm d} \in \mathbb{R}^{3 \times \rm{T_{d}}}\), the orientation of the end-effector as Euler angles \(\boldsymbol{\theta}_{\rm d} \in \mathbb{R}^{3 \times \rm{T_{d}}}\), the wrench exerted by the end-effector \(\mathbf{W}_{\rm d} \in \mathbb{R}^{6 \times \rm{T_{d}}}\), as well as the Cartesian velocities $\dot{\mathbf{X}}_{\rm d} \in \mathbb{R}^{3 \times \rm{T_{d}}}$ and angular end-effector velocities $\dot{\boldsymbol{\theta}}_{\rm d} \in \mathbb{R}^{3 \times \rm{T_{d}}}$.
In the proposed time-invariant, state-dependent approach the reference positions, velocities, and wrenches are determined using the current robot's end-effector position and orientation as\vspace{-0.1cm}
\begin{align}\label{eq:mg}
    \begin{pmatrix}
        \dot{\mathbf{x}}_{\rm ref}\\
        \dot{\boldsymbol{\theta}}_{\rm ref}\\
        \mathbf{w}_{\rm ref}
    \end{pmatrix} = f \left( \mathbf{x}(t), \boldsymbol{\theta}(t), \mathbf{x}_{\rm start}(t), \mathbf{x}_{\rm goal}(t), \mathbf{x}_{\rm obs}(t) \right).
\end{align}
Here, \(f\) is the dynamical system derived from the demonstration, which maps the current states $\mathbf{x}$ and $\boldsymbol{\theta}$ to the desired end-effector states $(\cdot)_{\rm ref}$. The derivation of $f$ is described in the following.
\par The encoded motion should be usable anywhere within the robot's workspace; therefore, we need to generalize it based on new starting and ending points. To adapt the demonstrated motion to time-varying start \(\mathbf{x}_{\rm start}(t)\in\mathbb{R}^{3}\) and goal \(\mathbf{x}_{\rm goal}(t)\in\mathbb{R}^{3}\) points, we employ scaling and rotation operations to transform the original demonstration. The scaling factor for translations is given by \vspace{-0.0cm} 
\begin{align} \label{eq:scale}
    s_{L}(t) = \frac{\| \mathbf{x}_{\rm goal}(t) - \mathbf{x}_{\rm start}(t)\|}{\| \mathbf{x}_{\rm d}^{\rm{T_{\rm d}}} - \mathbf{x}_{\rm d}^{1}\|},
\end{align}
where \(\ \mathbf{x}_{\rm d}^{\rm i}\in\mathbb{R}^{3}\) is the i-th sample point of end-effector position from demonstration $\mathbf{X}_{\rm d}$.

We compute the rotation matrix $\mathbf{R}(t) \in \mathbb{R}^{3 \times 3}$ using the Rodrigues' rotation formula \cite{DAI2015144}, which aligns the scaled 3D motion with the desired endpoint, i.e. the vectors $\mathbf{x}_{\rm d}^{\rm{T_{\rm d}}} - \mathbf{x}_{\rm d}^{1}$ and $\mathbf{x}_{\rm goal}(t) - \mathbf{x}_{\rm start}(t)$ are aligned.
From the scaling and rotation operation, the transformation of the position follows as \vspace{-0.2cm}
\begin{align}
    \Tilde{\mathbf{x}}_{\rm d}^{\rm t_{\rm d}}(t) &= s_{L}(t)\mathbf{R}(t) \Bigl( {\mathbf{x}}_{\rm d}^{{\rm t}_{\rm d}}- {\mathbf{x}}_{\rm d}^{1}\Bigr) +\mathbf{x}_{\rm start}(t) \quad \forall {\rm t}_{\rm d} \in \left[1,\rm{T_{\rm d}}\right].\label{eq:scaint} \nonumber\\[-0.2cm]
\end{align}
The linear velocities are scaled and rotated such that they are consistent with the transformed positions in (\ref{eq:scaint}). The end-effector orientation \(\boldsymbol{\theta}_{\rm d}(t)\) and angular velocities $\dot{\boldsymbol{\theta}}_{\rm d}$ are not transformed. 
Deviations of the end-effector from the desired trajectory due to external perturbations are compensated by feedback velocities that attract to the nearest point on that trajectory.
The index \( \mathrm{i_{\text{min}}}(t) \in \left[1,\rm{T_{\rm d}}\right]\) of the closest point from the demonstration to the current position is determined as \vspace{-0.6cm}
\begin{align}
    {\rm i}_{\text{min}}(t) = \arg\min_{{\rm i} \in \left[1,\rm{T_{\rm d}}\right]} \| \tilde{\mathbf{x}}_{\rm d}^{\rm i}(t) - \mathbf{x}(t) \|.
\end{align} 
The spatial \textit{kd}-tree algorithm can be employed to minimize execution time when finding the closest point \cite{ram2019revisiting}. The linear and angular feedback velocities are given by
\vspace{-0.2cm}
\begin{align}
    \mathbf{v}_{\rm {fb}}(t) = \boldsymbol{\Lambda}_{\rm L}(\Tilde{\mathbf{x}}_{\rm d}^{\rm i_{\rm{min}}}(t) - \mathbf{x}(t)),\nonumber\\
    \boldsymbol{\omega}_{\rm {fb}}(t) = \boldsymbol{\Lambda}_{\rm A}(\boldsymbol{\theta}_{\rm d}^{\rm i_{\rm{min}}}(t) - \boldsymbol{\theta}(t)).
\end{align}
Here $\boldsymbol{\Lambda}_{\rm L}\in\mathbb{R}^{3 \times 3}$ and $\boldsymbol{\Lambda}_{\rm A}\in\mathbb{R}^{3 \times 3}$ are positive definite feedback gains. These feedback velocities guide the end-effector toward the desired trajectory, but they do not guarantee tracking of the trajectory. Therefore, feed-forward velocities are introduced to aid tracking. The linear and angular feed-forward velocities are \vspace{-0.4cm}
\begin{align}
  \mathbf{v}_{\rm{ff}}(t) = \dot{\Tilde{\mathbf{x}}}_{\rm d}^{\rm i_{\rm{min}}}(t),\nonumber\\\boldsymbol{\omega}_{\rm ff}(t) = \dot{\boldsymbol{\theta}}_{\rm d}^{\rm i_{\rm{min}}}(t).  
\end{align}
In the absence of obstacles, feedback velocities contribute to the convergence of the end-effector toward the desired trajectories, while the feedforward velocities facilitate the motion to the desired goal point. However, during trajectory generation, avoiding obstacles is crucial. To ensure this, velocities to bypass obstacles are computed as \( \mathbf{v}_{\rm obs}(t) = \mathbf{v}_{\rm r}(t) + \mathbf{v}_{\rm g}(t),
\)
where \(\mathbf{v}_{\rm r}(t)\) are repulsion velocities from the obstacle, pushing the robot away from the obstacle, and \(\mathbf{v}_{\rm g}(t)\) are guidance velocities, which help to guide the robot during the obstacle avoidance. To simplify the calculations, obstacles in this paper are represented by bounding spheres.
We define \(\mathbf{n}_{\rm o}(t) \in \mathbb{R}^{3}\) as the normal vector to the obstacle bounding sphere \vspace{-0.4cm} 
\begin{align}
    \mathbf{n}_{\rm o}(t)= \frac{\mathbf{x}(t)-\mathbf{x}_{\rm obs}(t)}{\|\mathbf{x}(t)-\mathbf{x}_{\rm obs}(t)\|},\label{obsavoid}
\end{align}
where \(\mathbf{x}_{\rm obs}(t)\in\mathbb{R}^3\) is the centre of the obstacle.
The repulsion velocity is directed in the normal direction away from the surface of the bounding sphere 
\vspace{-0.2cm}
\begin{align}
    \mathbf{v}_{\rm r}(t) = \lambda_{\rm obs}(\mathbf{x},\mathbf{x}_{\rm obs}) \mathbf{n}_{\rm o}(t), \label{rpl}
\end{align}
where \(\lambda_{\rm obs}(\mathbf{x},\mathbf{x}_{\rm obs}) \in \mathbb{R}^{+} \) is a state-dependent positive gain that increases in the vicinity of the obstacle \vspace{-0.1cm}
\begin{align}
    \lambda_{\rm obs}(\mathbf{x},\mathbf{x}_{\rm obs}) &= \frac{1}{\left(\|\mathbf{x}(t)-\mathbf{x}_{\rm obs}(t)\| - r\right)}.
\end{align}
Here the radius of the obstacle is denoted as $r\in\mathbb{R}$.
The guidance velocities make it possible to give a preference for the obstacle avoidance movement by setting a desired avoidance direction 
\vspace{-0.2cm}
\begin{align}
    \mathbf{v}_{\rm g} (t) =  \lambda_{\rm obs}(\mathbf{x},\mathbf{x}_{\rm obs}) \mathbf{v}_{\rm proj}(t) k_g(t).
\end{align}
The vector $\mathbf{v}_{\rm proj}(t) \in \mathbb{R}^{3}$ consists of $\mathbf{v}_{\rm{ff}}(t)$ and a desired direction vector $\mathbf{v}_{\rm{dir}}$, which are projected onto the plane orthogonal to $\mathbf{n}_{\rm o}(t)$ \vspace{-0.2cm}
\begin{align}
    \mathbf{v}_{\rm proj}(t) &= \frac{\mathbf{v}_{\rm{ff}}(t)}{\|\mathbf{v}_{\rm{ff}}(t)\|} - \left\langle \frac{\mathbf{v}_{\rm{ff}}(t)}{\|\mathbf{v}_{\rm{ff}}(t)\|},\mathbf{n}_{\rm o}(t)\right\rangle \mathbf{n}_{\rm o}(t) \\ \nonumber
    & + \mathbf{v}_{\rm{dir}} - \left\langle \mathbf{v}_{\rm{dir}},\mathbf{n}_{\rm o}(t)\right\rangle \mathbf{n}_{\rm o}(t).
\end{align}
The dot product of two vectors is denoted as $\left\langle \cdot,\cdot \right\rangle$.
The desired direction vector $\mathbf{v}_{\rm{dir}}$ is selected manually, e.g. in the direction of the z-axis of the world frame to avoid the obstacle by moving over it. 
The guidance gain $k_g(t)\in \{0,1\}$ activates the guidance velocity if the angle between the normal vector to the obstacle bounding sphere and the feed-forward velocity is bigger than $\ang{90}$
\vspace{-0.1cm}
\begin{align}
    k_g(t) = 
    \begin{cases} 
    1, & \text{if } \left\langle \mathbf{v}_{\rm{ff}}(t),\mathbf{n}_{\rm o}(t)\right\rangle \leq 0, \\
    0, & \text{otherwise}.
    \end{cases}
\end{align}
If the angle between the vectors is less than $\ang{90}$, the system effectively navigates past obstacles and onto the desired path by combining feed-forward, feedback, and obstacle-repulsion velocities. This is facilitated by the positive feedback mechanism described by (\ref{obsavoid}).

\par Further, velocity modulation is crucial for harmonizing manipulator behavior with varying levels of human autonomy, thereby facilitating a smooth transition of authority and ensuring that the generated reference trajectory remains within bounds.
To address this, we modulate the motion generator and consider system dynamics as \vspace{-0.2cm}
\begin{align}
    \dot{\hat{\mathbf{x}}}_{\rm ref}(t) &= (1-\alpha_{\rm h}(t))^2\biggl(\mathbf{v}_{\rm fb}(t) + \mathbf{v}_{\rm ff}(t) + \mathbf{v}_{\rm obs}(t) \biggr), \label{eq:final_vel}
\end{align}
here \(\alpha_{\rm h}(t) \in [0,1]\) is the human authority value, where \(\alpha_{\rm h} = 1\) signifies the human operator having a leadership role in the collaborative task\textcolor{black}{, see section \ref{sec:aa}}. \textcolor{black}{Velocities are scaled using a quadratic reduction factor of authority value instead of a linear one to ensure faster decay of reference velocity.} The norm of the velocity is limited to a manually selected value $v_{\rm th} \in \mathbb{R}^{+}$ by
\vspace{-0.15cm}
\begin{align}
     \dot{\mathbf{x}}_{\rm ref}(t) =
\begin{cases}
     \dot{\hat{\mathbf{x}}}_{\rm ref}(t), & \text{if } \|  \dot{\hat{\mathbf{x}}}_{\rm ref}(t) \| \leq v_{\rm th} \\
      \dot{\hat{\mathbf{x}}}_{\rm ref}(t) \frac{v_{\rm th}}{\|  \dot{\hat{\mathbf{x}}}_{\rm ref}(t) \|}, & \text{otherwise}.
\end{cases}\label{sigma}
\end{align}
(\ref{eq:final_vel}) and (\ref{sigma}) guarantee that the generated reference velocities are bounded, thereby enhancing system safety. 
\textcolor{black}{The combination of feedback, feedforward, and obstacle avoidance velocities creates the virtual potential field in order to effectuate the desired motion with attraction to a reference trajectory, following the reference trajectory with its velocity profile while also avoiding detected obstacles.}

\par Further outputs from the dynamical system in (\ref{eq:mg}) are the reference angular velocities and wrenches. The reference angular velocities are determined as $\dot{\boldsymbol{\theta}}_{\rm ref}(t) = (1-\alpha_{\rm h}(t))^2 \dot{\boldsymbol{\theta}}_{\rm d}^{\rm i_{\text{min}}}(t)$. We define the desired wrench \(\mathbf{w}_{\rm ref}(t) \in \mathbb{R}^{6}\) as \vspace{-0.5cm}
\begin{align}
\mathbf{w}_{\rm ref}(t) = \beta(t)  \mathbf{W}_{\text{d}}^{\rm i_{\text{min}}}(t),
\end{align}
where \(\beta(t) \in \{0,1\}\) is given by
\vspace{-0.1cm}
\begin{align}
    \resizebox{\linewidth}{!}{
    $\beta(t) = 
    \begin{cases} 
        (1 - \alpha_{\rm h}(t))^2, & \text{if } \left\| \tilde{\mathbf{X}}_{\text{d}}^{i_{\text{min}}}(t) - \mathbf{x}(t) \right\| + \left\| \tilde{\boldsymbol{\theta}}_{\text{d}}^{i_{\text{min}}}(t) - \boldsymbol{\theta}(t) \right\| \leq w_{\text{th}}, \\
        0, & \text{otherwise}.
    \end{cases}$
    }\nonumber\\[-0.3cm]
\end{align}
Here, \(\beta(t)\) ensures that the robot generates force only when the position errors are below the threshold \(w_{\text{th}} \in \mathbb{R}^{+}\) and the robot has autonomy, i.e. $\alpha_{\rm h}$ is low; otherwise, it remains zero. The focus now transitions to the computation of authority allocation.

\subsection{Authority allocation between human and robot}\label{sec:aa}
The authority allocation between the human operator and the robot allows the human to take over control of the robot by applying forces to the robot.
To facilitate a smooth transition of control between the tracking controller and the human operator, we introduce a variable stiffness \(\mathbf{K}(t) \in \mathbb{R}^{6\times6}\) and a variable damping \(\mathbf{D}(t) \in \mathbb{R}^{6\times6}\), both of which are functions of the human authority parameter \(\alpha_{\rm h}(t)\). The variable stiffness is defined as \(\mathbf{K}(t) = (1 - \alpha_{\rm h}(t)) \mathbf{K}_{\rm max}\).
Subsequently, the variable damping \textcolor{black}{is} selected as $\mathbf{D}(t) = 2 \sqrt{\mathbf{K}(t)}$. 

\textcolor{black}{The human authority parameter \(\alpha_{\rm h}(t)\) is adjusted based on external perturbations from humans or the environment
\vspace{-0.2cm}
\begin{align}\label{eq:alpha^}
    \hat{\alpha}_{\rm h}(t) = 0.5\left(1+\tanh\left(\frac{3}{a} \left({\rm w}_{\rm diff} - a - b\right)\right)\right).
\end{align}
\begin{figure}[!htbp]
  \centering
\includegraphics[width=0.8\linewidth,keepaspectratio]{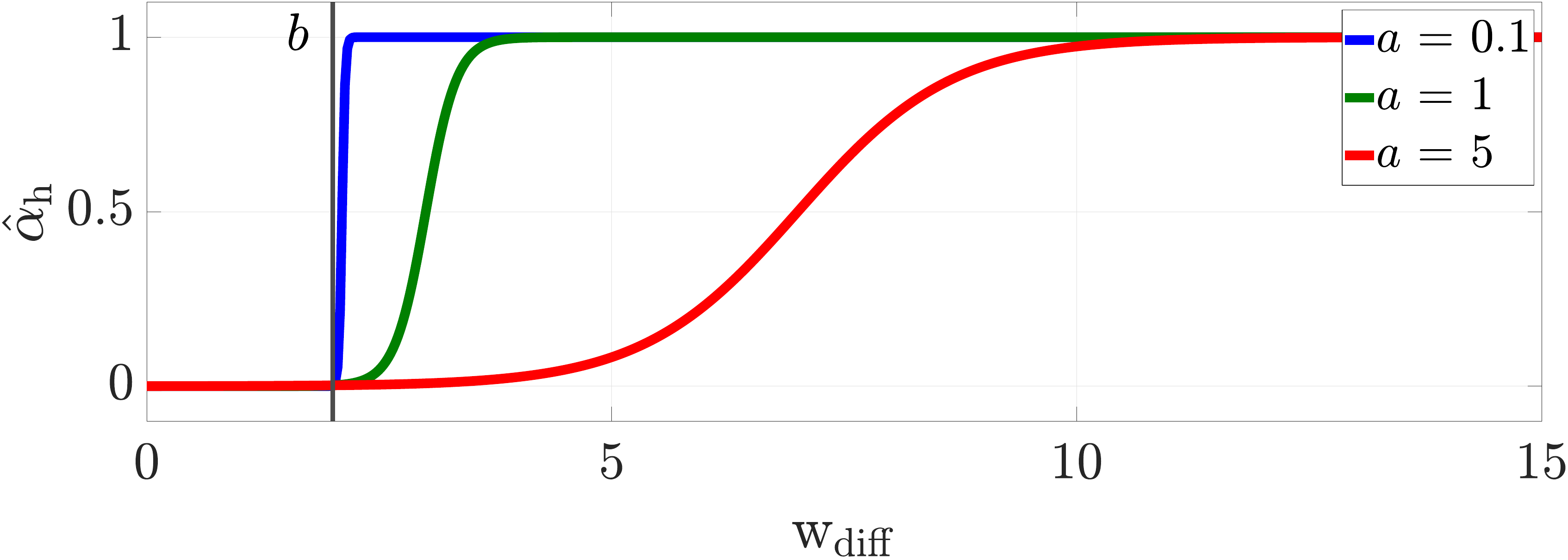}
  \vspace{-0.3cm}
  \caption{
  Parameter influence on \( \hat{\alpha}_{\text{h}} \) calculation with \( b = 2 \) setting the deadband for \( \hat{\alpha}_{\text{h}} \), and \( a \) setting the slope of the \( \hat{\alpha}_{\text{h}} \) transition to the maximum of 1.}
  \label{fig:alpha_update}
  \vspace{-0.8cm}
\end{figure}
Here, \(a > 0\) modulates the slope of the transition function and \(b > 0\) determines the width of the deadband.
The perturbations are considered as the norm of wrench difference ${\rm w}_{\rm diff} = c_1\|\mathbf{w}_{\rm s}(t) - \mathbf{w}_{\rm est}(t)\| + c_2\|\mathbf{w}_{\rm ref}(t) - \mathbf{w}_{\rm s}(t)\|$ between the reference wrench $\mathbf{w}_{\rm ref}(t)$, the measured wrench \(\mathbf{w}_{\rm s}(t) \in \mathbb{R}^{6}\) from a force-torque sensor, and the wrench \(\mathbf{w}_{\rm est}(t) \in \mathbb{R}^{6}\) estimated based on joint torque measurements. The wrench difference ${\rm w}_{\rm diff} \in \mathbb{R}^{+}$ is separated into two parts with respective scaling factors \(c_1 \& c_2 > 0\). The first part of ${\rm w}_{\rm diff}$ allows for the integration of wrenches applied to the robot that are not captured by the sensor while the second part considers the error of the measured wrench \(\mathbf{w}_{\rm s}(t)\) compared to the reference wrench $\mathbf{w}_{\rm ref}(t)$.
\textcolor{black}{This two-part solution not only enables the user to touch anywhere on the robot to take over the authority but also allows for the lowering of the robot's authority, thereby increasing its compliance when wrenches are applied due to accidental contact or human intervention.} Finetuning the sensitivity of $\hat{\alpha}_{\rm h}(t)$ to different kinds of external forces is achievable through the two scaling factors, e.g. making the authority allocation less responsive to force errors applied tool-side of the sensor by selecting $c_1 > c_2$, which can enhance the force production capabilities.} \textcolor{black}{The parameters $a$, $b$, $c_1$, and $c_2$ are empirically chosen design parameters of the authority arbitrator.}

A simplified illustration of this function and its parameters is shown in Fig. \ref{fig:alpha_update}. At every time step \(\alpha_{\rm h}(t)\) is updated based on the computed $\hat{\alpha}_{\rm h}(t)$ value using a recursive filter formulation, which facilitates a smoother transition of $\alpha_{\rm h}$ values
\vspace{-0.4cm}
\begin{align}
    \alpha_{\rm h, {i}} = \alpha_{\rm h, {i-1}} + g_{\alpha} \left( \hat{\alpha}_{\rm h} - \alpha_{\rm h, {i-1}} \right),
\end{align}
where $\left(\cdot\right)_{\rm i}$ denotes the i-th time step and $g_{\alpha}(t) \in\left[0,1\right]$ is the update gain \vspace{-0.2cm}
\begin{align} \label{eq:alpha_filt_gain}
    g_{\alpha}(t) = 
    \begin{cases} 
    g^{\rm +}, & \text{if } \hat{\alpha}_{\rm h} > \alpha_{\rm h, {i-1}}, \\
    g^{\rm -} + g^{\rm -} \left( 1 - \alpha_{\rm h, {i-1}} \right) ^2 , & \text{otherwise}.
    \end{cases}
\end{align}
By setting $g^{\rm +}$ higher than $g^{\rm -}$ a faster increase rate than decrease rate of $\alpha_{\rm h}$ can be achieved.
These parameters as well as the previously mentioned $a$, $b$, $c_1$, and $c_2$ values were manually tuned when implementing the approach on the robot.
The time-varying impedance parameters may induce instabilities in the closed-loop system therefore we design our closed-loop controller using the energy tank approach of passivity theory \cite{energyTankFed,schindlbeck2015unified}.
\subsection{Variable Impedance and Force Controller}
We consider an n degrees of freedom manipulator's interaction in task-space \cite{ott2008cartesian}, where the dynamics are characterized by
\vspace{-0.1cm}
\begin{align}
   \mathbf{M}_{\rm x}(\mathbf{q})\ddot{\boldsymbol{\xi}}(t) + \mathbf{C}_{\rm x}(\mathbf{q},\dot{\mathbf{q}},{\mathbf{v}_n})\dot{\boldsymbol{\xi}}(t) + \mathbf{g}_{\rm x}(\mathbf{q}) &= \mathbf{w}(t) + \mathbf{w}_{\mathrm{env}}(t). \label{eq:robot_dynamics}
\end{align}
Here, \( \mathbf{q}(t) \in \mathbb{R}^n \) is the vector of joint angles, \(\boldsymbol{\xi}(t)\in \mathbb{R}^{6}\) is the Cartesian pose of the manipulator, \({\mathbf{v}_n}\) are null-space velocities, \(\mathbf{M}_{\rm x}(\mathbf{q}) \in \mathbb{R}^{6 \times 6} \) is the inertia matrix, \(\mathbf{C}_{\rm x}(\mathbf{q},\dot{\mathbf{q}}) \in \mathbb{R}^{6 \times 6} \) represents the Coriolis and centripetal forces, \(\mathbf{g}_{\rm x}(\mathbf{q}) \in \mathbb{R}^6 \) is the vector of conservative wrench, \(\mathbf{w}(t) \in \mathbb{R}^6 \) is the control input to the manipulator, and \(\mathbf{w}_{\mathrm{env}}(t) \in \mathbb{R}^6 \) represents wrench from interactions with the environment. We define the control law as \vspace{-0.15cm}
\begin{align} \label{eq:control_law}
    \mathbf{w}(t) &= \mathbf{M}_{\rm x}(\mathbf{q})\ddot{\boldsymbol{\xi}}_{\rm ref}(t) + \mathbf{C}_{\rm x}(\mathbf{q},\dot{\mathbf{q}},{\mathbf{v}_n})\dot{\boldsymbol{\xi}}_{\rm ref}(t)  + \mathbf{g}_{\rm x}(\mathbf{q})  \\\nonumber & - \mathbf{\bar{K}}\mathbf{e}(t) - \mathbf{\bar{D}}\dot{\mathbf{e}}(t) + \mathbf{u}(t), 
\end{align}
where \(\mathbf{\bar{K}}\in\mathbb{R}^{6\times 6}\) is positive definite stiffness, \(\mathbf{\bar{D}}\in\mathbb{R}^{6\times 6}\) is positive definite  damping, \(\mathbf{e}(t) = \boldsymbol{\xi}(t) - \boldsymbol{\xi}_{\rm ref}(t)\) is the tracking error and \(\mathbf{u}(t)\in\mathbb{R}^{6}\) is an additional control input with time-varying impedance and wrench parameters. The motion generator, governed by (\ref{eq:final_vel}), generates desired reference trajectories utilizing feedback from the manipulator's current states. In the absence of the control input \(\mathbf{u}(t)\), the robot demonstrates high compliance to human-induced perturbations and deviates from the reference trajectory and wrench, a result of choosing small values for \(\mathbf{\bar{K}}\) and setting \(\mathbf{\bar{D}} = 2\sqrt{\mathbf{\bar{K}}}\). Conversely, \(\mathbf{u}(t)\) enables the integration of the authority level for trajectory and force tracking with variable impedance, as delineated in Section \ref{sec:aa}. The control input is inspired by \cite{energyTankFed,schindlbeck2015unified,ENAYATI2020} and defined as\vspace{-0.15cm}
\begin{align}\label{tvi}
    &\mathbf{u}(t) = \zeta \Bigl(  -\mathbf{K}(t)\mathbf{e} - \mathbf{D}(t)\dot{\mathbf{e}}  + (1-\phi)\mathbf{w}_{\rm ref}(t)\nonumber\\ &+ \mathbf{K}_{\rm w}(\mathbf{w}_{\rm env}(t) - \mathbf{w}_{\rm ref}(t))\nonumber\\ &+ \mathbf{K}_{\rm i}\int_{0}^{t}(\mathbf{w}_{\rm env}(\delta) - \mathbf{w}_{\rm ref}(\delta)) d\delta \Bigr)+ \phi\mathbf{w}_{\rm ref}(t)
\end{align}
where \(\mathbf{K}(t) = (1 - \alpha_{\rm h}(t)) \mathbf{K}_{\rm max}\) is variable stiffness, \(\mathbf{D}(t) = 2 \sqrt{\mathbf{K}(t)}\) is variable damping, and \(\mathbf{K}_{\rm i} \hspace{0.1cm}\& \hspace{0.1cm}\mathbf{K}_{\rm  w}\in\mathbb{R}^{6 \times 6}\) are positive-definite proportional and integral gain for wrench error. \(\zeta\in\{0,1\}\) and \(\phi\in\{0,1\}\) are associated with the energy tank, which will be explained in more detail in the following. The tank energy is defined as
\(\psi_{\mathrm{energy}}(t) = \frac{1}{2}s^2(t)\), where \(s\in\mathbb{R}\) is the state associated with energy-tank.
For passivity analysis, we consider the storage function as
\vspace{-0.1cm}
\begin{align}
    v = \frac{1}{2}\dot{\mathbf{e}}^{\top}\mathbf{M}_{\rm x}\dot{\mathbf{e}} +\frac{1}{2}{\mathbf{e}}^{\intercal}\bar{\mathbf{K}}{\mathbf{e}}  +\frac{1}{2}s^2.
\end{align}
Utilizing the skew-symmetric property \(\frac{1}{2}\dot{\mathbf{e}}^{\intercal}(\dot{\mathbf{M}}_{x} - 2 \mathbf{C}_{x})\dot{\mathbf{e}}=0\), the derivative \(\dot{v}\) becomes
\vspace{-0.15cm}
\begin{align}
    \dot{v}= -\dot{\mathbf{e}}^{\intercal}\bar{\mathbf{D}}\dot{\mathbf{e}}+ \dot{\mathbf{e}}^{\intercal}\mathbf{w}_{\mathrm{env}} + \dot{\mathbf{e}}^{\intercal}\mathbf{u} +s\dot{s}.\label{vdot1}
\end{align}

Herein, the terms \(\dot{\mathbf{e}}^{\intercal}\mathbf{w}_{\mathrm{env}}\) and \(\dot{\mathbf{e}}^{\intercal}\mathbf{u}\) are observed to be sign-indefinite. By invoking Equation (\ref{tvi}), \(\dot{s}\) is formulated as
\vspace{-0.3cm}
\begin{align}
    \label{sdot}\dot{s}&=\frac{\gamma}{s}(\dot{\mathbf{e}}^{\intercal}\bar{\mathbf{D}}\dot{\mathbf{e}} - \phi\dot{\mathbf{e}}^{\intercal}\mathbf{w}_{\rm ref}(t) ) - \frac{\zeta}{s}\biggl(-\dot{\mathbf{e}}^{\intercal}\mathbf{K}(t)\mathbf{e} - \dot{\mathbf{e}}^{\intercal}\mathbf{D}(t)\dot{\mathbf{e}} \nonumber\\ &+ (1-\phi)\dot{\mathbf{e}}^{\intercal}\mathbf{w}_{\rm ref}(t)\nonumber+\dot{\mathbf{e}}^{\intercal}\mathbf{K}_{\rm w}(\mathbf{w}_{\rm env}(t) - \mathbf{w}_{\rm ref}(t)) \nonumber\\& +\dot{\mathbf{e}}^{\intercal}\mathbf{K}_{\rm i}\int_{0}^{t}(\mathbf{w}_{\rm env}(\delta) - \mathbf{w}_{\rm ref}(\delta)) d\delta\biggr),
\end{align}
where \vspace{-0.5cm}
\begin{align}
    \gamma(t) =
\begin{cases}
    1, & \text{if } \psi_{\mathrm{energy}}(t) \leq \overline{\psi}_{\mathrm{threshold}} \\
    0, & \text{otherwise}
\end{cases}
\end{align}
with \(\overline{\psi}_{\mathrm{threshold}}\in\mathbb{R}^{+}\) serving as an application-dependent upper bound to preclude excessive energy accumulation in the reservoir.
\begin{figure}[!htbp]
  \centering
\includegraphics[width=\linewidth,keepaspectratio]{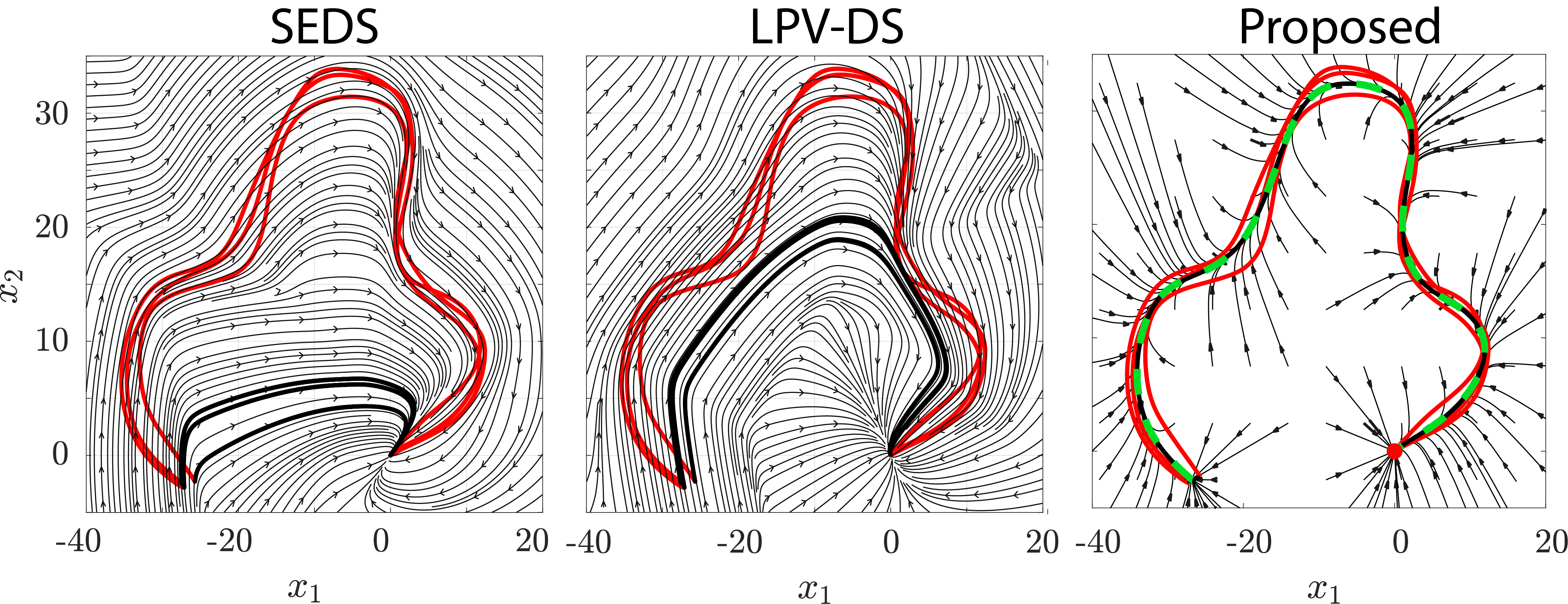}
  \vspace{-0.7cm}
  \caption{Comparison of the Proposed Algorithm with SEDS and LPV-DS on the LASA handwriting dataset.}
  \label{fg:comp}
  \vspace{-0.5cm}
\end{figure}
To avert singularities, it is required that \(s(0)>0\). The parameter \(\zeta\), controlling the extraction of energy due to non-passive actions, is defined as
\vspace{-0.5cm}
\begin{align}
    \zeta(t) =
\begin{cases}
    1, & \text{if } \psi_{\mathrm{energy}}(t) \geq \underline{\psi}_{\mathrm{threshold}} \\
    0, & \text{otherwise},
\end{cases}
\end{align}
where \(\underline{\psi}_{\mathrm{threshold}} \in \mathbb{R}^{+}\) constitutes a positive lower energy boundary to ensure that the energy tank is not fully depleted. Additionally, we use dissipative reference wrench to quickly refill the energy tank. The parameter \(\phi\), which modulates desired wrench effects, is set as follows:
\vspace{-0.1cm}
\begin{align} \label{eq:phi}
    \phi(t) =
\begin{cases}
    1, & \text{if } \dot{e}^{\intercal}\mathbf{w}_{\text{ref}}(t)< 0 \\
    0, & \text{otherwise},
\end{cases}
\end{align}
This setting allows the controller to keep applying feed-forward wrench even if the energy tank is empty, as shown in (\ref{tvi}). Substituting (\ref{sdot}) in (\ref{vdot1}) and as \(\gamma,\zeta,\phi\in\{0,1\}\) we get \(\dot{v}\leq \dot{e}^{\intercal}\mathbf{w}_{\mathrm{env}}\), which is the  condition required for passivity for pair (\(\dot{e},\mathbf{w}_{\mathrm{env}}\)).
\textcolor{black}{The parameters of the variable impedance controller and the energy tank are empirically chosen.}

\section{Simulations and Experiments}

To assess the efficacy of the proposed methodology, we conducted a comparative evaluation against SEDS and LPV-DS, utilizing the leaf shape from the LASA handwriting dataset, as depicted in Fig. \ref{fg:comp}. While both SEDS and LPV-DS succeeded in reaching the endpoint, they failed to accurately replicate the original demonstration. Although LPV-DS exhibited lower tracking error compared to SEDS, it did not faithfully reproduce the original demonstration.

\begin{figure}[!hbtp]
\centering
\includegraphics[width=0.6\linewidth,keepaspectratio]{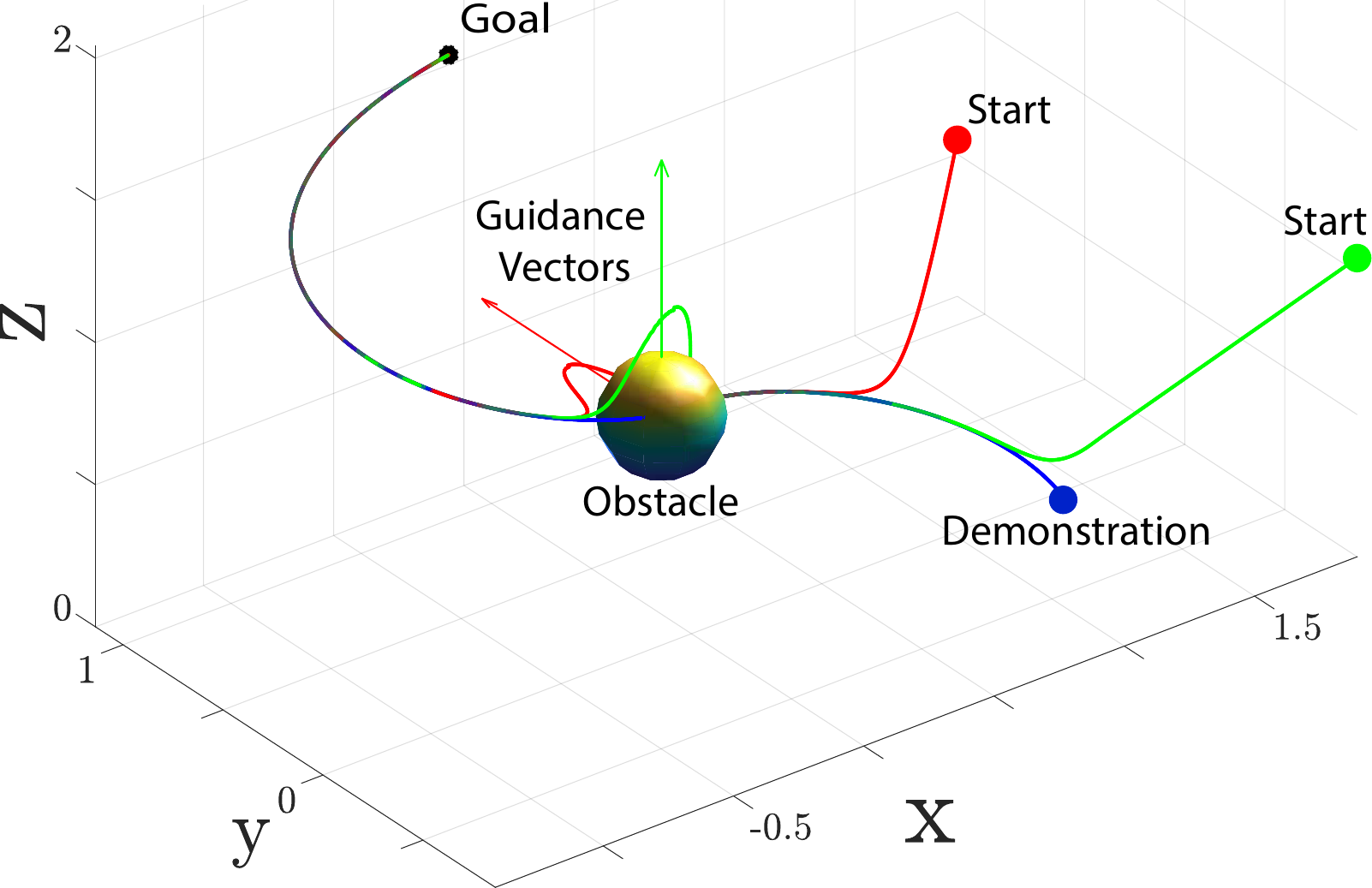}
  \caption{Demonstration of obstacle avoidance using the proposed approach, following the desired guidance direction.}
  \label{fg:3d_salad}
  \vspace{-0.5cm}
\end{figure}

In Fig. \ref{fg:comp}., the red lines represent the original demonstrations, the black lines with arrows indicate the streamline of the velocity field, and the solid black lines denote the reproduced demonstrations. For our proposed approach, we employed the mean trajectory of the demonstrations, highlighted by the green dashed line in Fig. \ref{fg:comp}. However, it is important to note that the proposed methodology necessitates only a single demonstration, selected from multiple demonstrations based on optimal performance criteria. Our results demonstrate that points originating away from the demonstration converge to and subsequently follow this mean trajectory, ultimately reaching the goal. This validates the capability of the proposed approach to encode complex shapes, that do not monotonically decrease in distance to the target over time. Similarly the proposed algorithm can be successfully deployed on the other shapes of the LASA handwriting dataset.
Additionally, we model the obstacle avoidance scenario, as depicted in Fig. \ref{fg:3d_salad}. In this case, the guidance vectors align with the positive y and z axes, directing the desired trajectories around the obstacle and ultimately converging to the goal point. 

\par The efficacy of the proposed approach is further validated through two distinct experimental studies. The first experiment focuses on a button-pressing task, highlighting the approach's proficiency in precise force generation and responsiveness to human perturbations. The second experiment evaluates the robot's capability to replicate complex movements on the real robot, akin to those depicted in Fig. \ref{fg:comp}, while ensuring safety during collaborative task execution. Real-time obstacle avoidance is also verified in the second experiment. A Franka Emika Research 3 robotic arm is employed in both studies. We use a 6-axis force-torque sensor from AIDIN ROBOTICS Inc. mounted at the robot's end-effector to measure interactions with the environment.
\subsection{Force production and reactivity to human perturbation}
\begin{figure}[!htbp]
  \centering
  \includegraphics[width=0.6\linewidth,keepaspectratio]{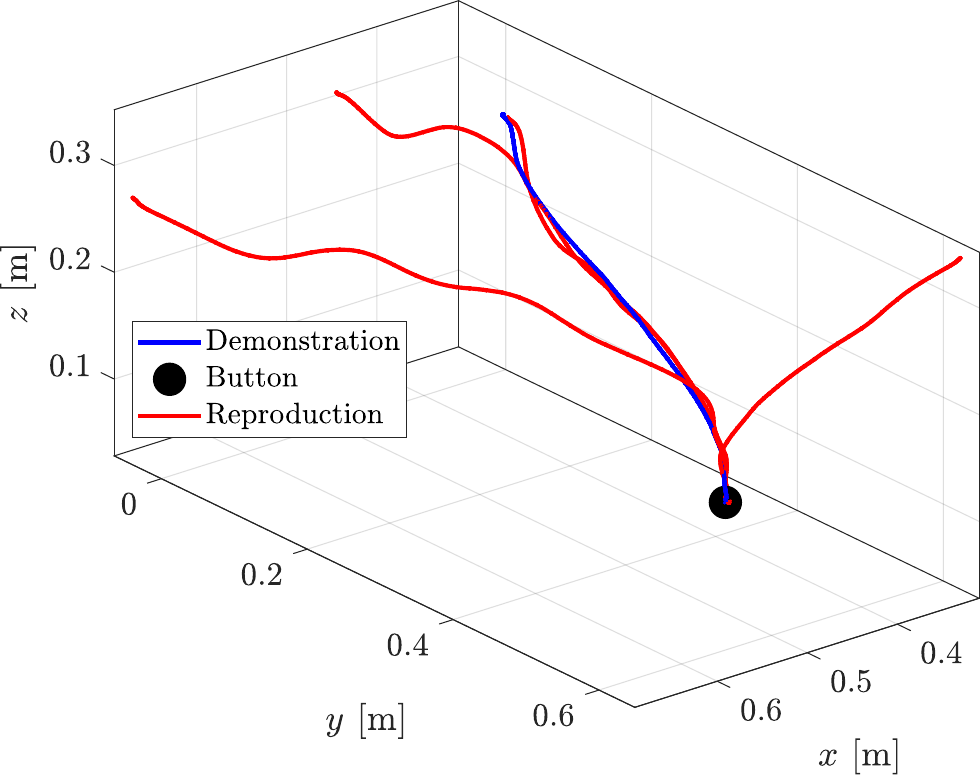}
  \vspace{-0.3cm}
  \caption{3D paths of the robot's end-effector position from demonstration (blue) and reproductions (red) of a button pressing task.}
  \label{fig:3D_Button}
  \vspace{-0.7cm}
\end{figure}

As described earlier, many tasks performed in contact with the environment require accurate force production. In this experiment, we investigate the ability of the proposed approach to reproduce the motion and interaction forces required to perform a task that is taught using kinesthetic teaching. Furthermore, it is important for the human to be able to take control of the robot during the collaboration and that the robot's motion is subsequently adapted. To demonstrate these characteristics, we chose the scenario of pressing a button where the human diverts the robot from the desired motion by repeatedly taking over autonomy, as illustrated in Fig. \ref{fg:setup} (right).
The human moves the robot's end-effector to different positions, from which the robot takes back control when the human stops applying forces. Thereafter the end-effector moves according to the previously defined attraction and feed-forward velocities, and finally converges to the target point (button) and presses it. Fig. \ref{fig:3D_Button} shows the 3D path of the end-effector during the demonstration and the reproduction of the motion using the proposed approach.

\begin{figure}[!htbp]
  \centering
  \vspace{0cm}
\includegraphics[width=0.8\linewidth,keepaspectratio]{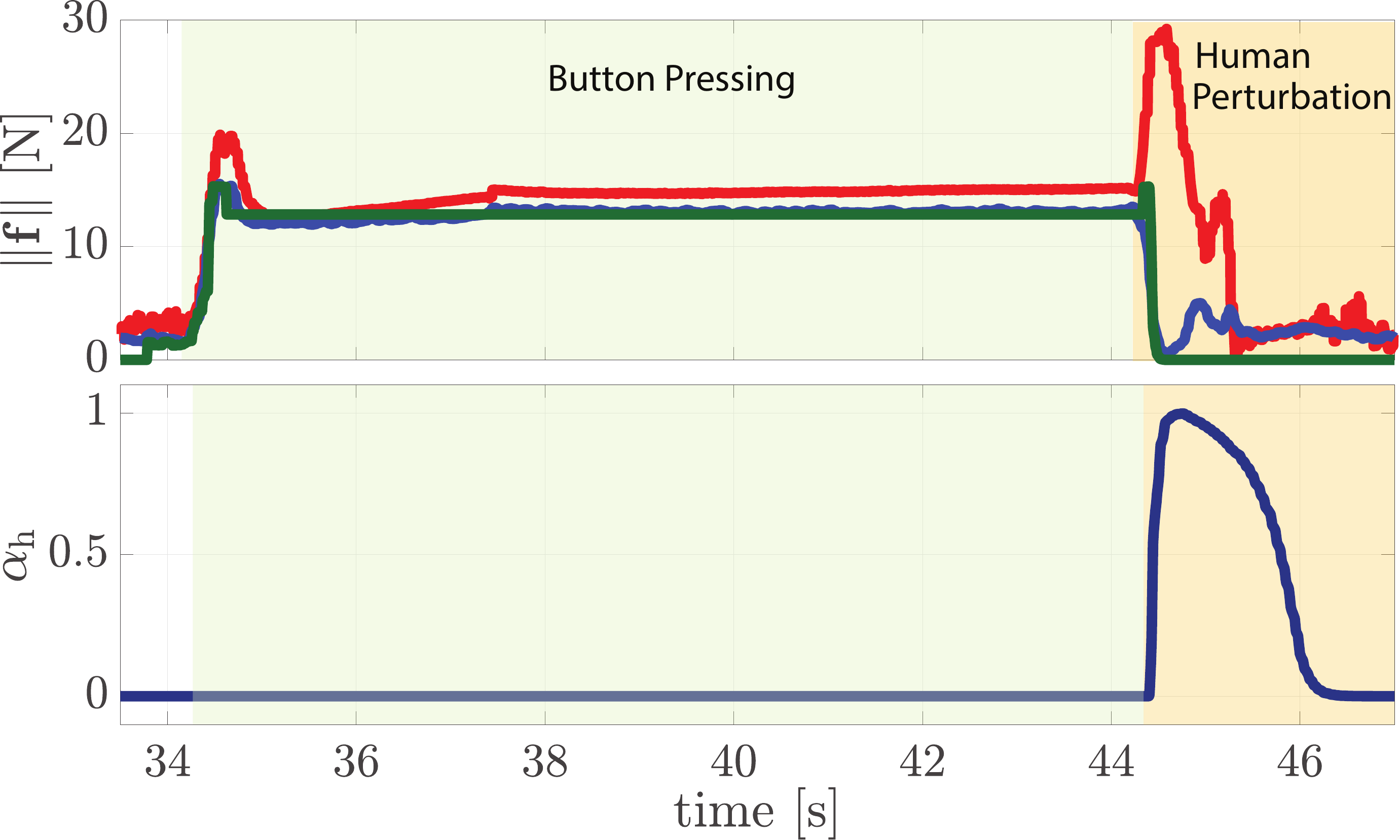}
  \vspace{-0.4cm}
  \caption{Reproduction of the button pressing task where the human uses shared autonomy to reposition the end-effector. The autonomy value $\alpha_{\rm h}$ (orange) is calculated based on the desired $\mathbf{f}_{\rm ref}$ (green), the measured $\mathbf{f}_{\rm s}$ (blue), and the estimated $\mathbf{f}_{\rm est}$ (red) end-effector forces.}
  \label{fig:alpha}
\end{figure}

By applying forces the human increases the authority allocation value according to  (\ref{eq:alpha^}) - (\ref{eq:alpha_filt_gain}), which leads to the robot assuming a more compliant role. Fig. \ref{fig:alpha} shows the absolute desired, measured, and estimated end-effector forces, as well as the human autonomy value $\alpha_{\rm h}(t)$. The emergency stop button chosen for this task requires approximately 15 $\si{\newton}$ to be pressed, which is encoded during the demonstration. 
As the measured force in Fig. \ref{fig:alpha} shows, the force required to trigger the button is reached during reproduction of the task. After the button is pressed (starting at 34.3 $\si{\second}$ in Fig. \ref{fig:alpha}), the reference force decreases slightly and the controller starts tracking the new desired reference force. During the force production period the maximum force error between desired and measured force in the direction of button press of 1.313 $\si{\newton}$ was observed. After approximately 2.5 $\si{\second}$ the force error reduces to a steady state tracking error of 0.142 $\si{\newton}$ due to the effects of the integral term in (\ref{tvi}). Despite the control parameter not being optimized for this specific task prior to the experiments, this highlights the force production capabilities of the proposed approach. Furthermore Fig. \ref{fig:alpha} shows starting at 44.2 $\si{\second}$ exemplary how the human applies forces to the robot's body (flange side of the FT sensor) is detected by the force estimation. Due to the mismatch of FT sensor measured forces and estimated forces, $\alpha_{\rm h}(t)$ increases with an approximate delay of approximately 150 $\si{\ms}$. The human can successfully take over control of the robot, which transitions to lower impedance. \vspace{-0.2cm}

\subsection{Encoding of complex closed-loop trajectories}
\begin{figure}[!htbp]
  \centering
   \vspace{-0.43cm}
\includegraphics[width=0.8\linewidth,keepaspectratio]{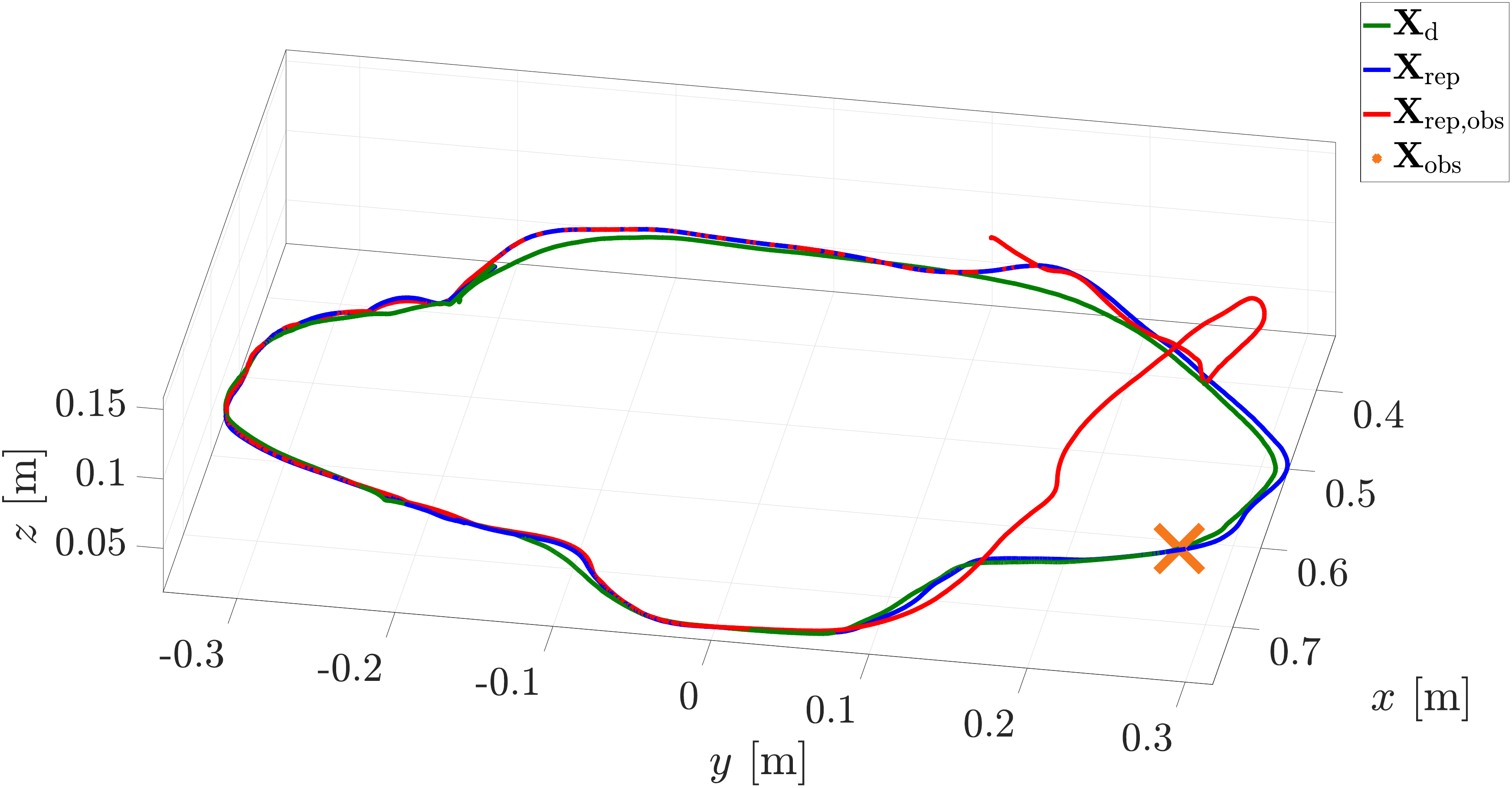}
  \vspace{-0.3cm}
  \caption{3D paths of the robot's end-effector position from demonstration (green), reproduction (blue), and reproduction while avoiding an obstacle (red).}
  \label{fig:3D_polishing}
  \vspace{-0.3cm}
\end{figure}

In the previous experiment, we already show that with the proposed approach the human is capable of taking over autonomy over the task by applying force on the robot. Therefore, in this experiment, we highlight the framework's ability to encode complex and closed-loop motion, avoid detected obstacles, and illustrate the workings of the energy tank. The task at hand deals with the polishing of an object with a complex shape, see Fig. \ref{fg:setup} (left). The shape encoded in this experiment is similar to the leaf shape of the LASA handwriting dataset, shown earlier in Fig. \ref{fg:comp}. The robot's end-effector position during demonstration, reproduction, and obstacle avoidance are shown in Fig. \ref{fig:3D_polishing}. The encoded, counter-clockwise motion can be reproduced in a closed-loop manner with the proposed approach. 
\begin{figure}[!htbp]
  \centering
  \vspace{-0.3cm}
  \includegraphics[width=0.7\linewidth,keepaspectratio]{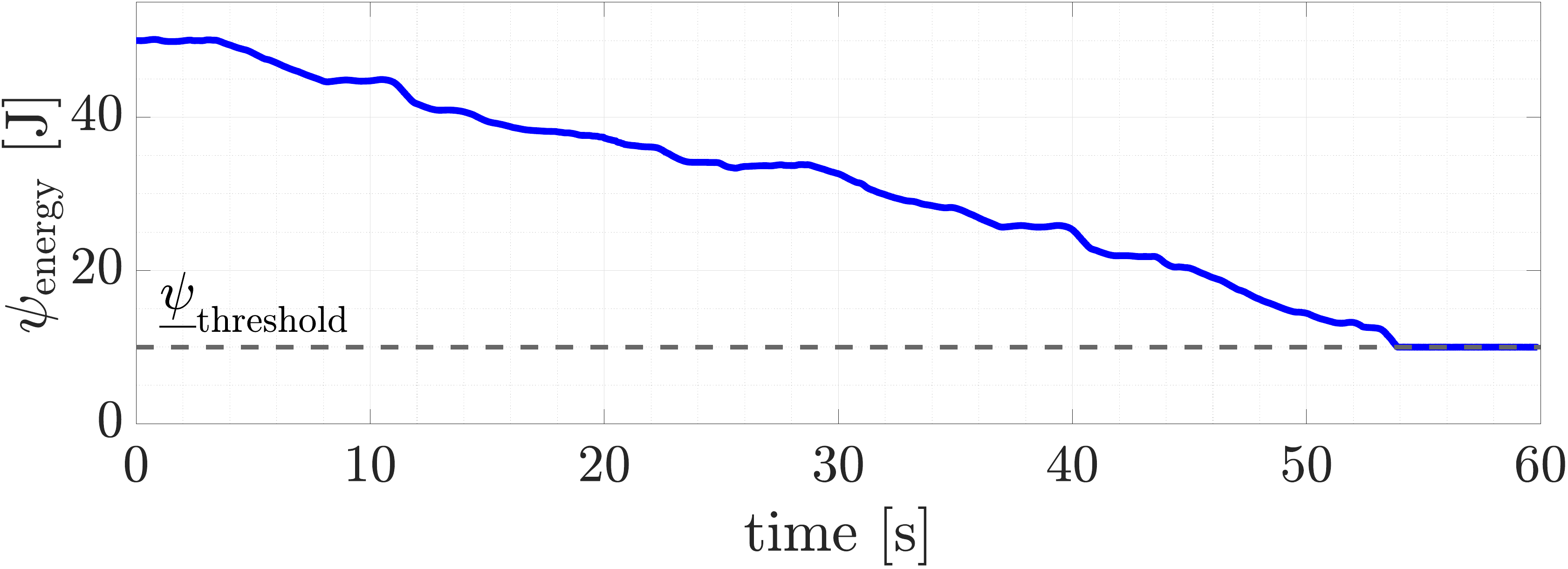}
  \vspace{-0.3cm}
  \caption{Tank energy $\psi_{\mathrm{energy}}$ during human robot collaboration experiment. The energy level decreases until the lower tank energy threshold $\underline{\psi}_{\mathrm{threshold}} = 10\ \si{\joule}$ is reached.}
  \label{fig:tank}
  \vspace{-0.8cm}
\end{figure}
  Additionally, in case of a detected obstacle, the motion generator is capable of avoiding the obstacle, passing it, and transitioning back to the reproduction of the task. Here ArUco markers \cite{GARRIDOJURADO20142280} placed on the object were used for obstacle detection \textcolor{black}{and the obstacle radius was selected empirically}. When the energy tank reaches its lower threshold, the robot switches to low impedance mode and ceases to track the desired motion, a state also achieved at the end of obstacle-avoiding motion.

As shown in  (\ref{eq:control_law}) - (\ref{eq:phi}), the implemented energy tank-based variable impedance control is able to ensure passivity of the controller. Fig. \ref{fig:tank} illustrates how the tank energy depletes over time while reproducing the encoded motion of this experiment.

At 55 $\si{\second}$ the energy in the tank reaches the lower limit and the respective parts of the controller, i.e. variable stiffness and damping term as well as the non-dissipative terms responsible for force production, are turned off. The robot is in a state of low impedance.

\section{Conclusion}
This study introduces a framework for shared autonomy, incorporating several key components. Initially, a time-invariant, state-dependent motion generator based on potential fields is presented. This approach encodes both complex and closed-loop task-specific trajectories and wrench profiles, while concurrently facilitating obstacle avoidance. Subsequently, the framework provides a methodology for dynamic and smooth authority transitions between human operators and robotic systems, employing variable impedance and force control mechanisms to ensure precise task execution and reactiveness to human-induced perturbations. Additionally, the framework integrates an energy-tank-based passivation strategy to maintain system passivity under time-varying parameters. The efficacy of the proposed framework is validated through experimental and simulation-based evaluations. Future research endeavors will concentrate on the incremental refinement of robotic skills in dynamic environments and the predictive modeling of human intentions to optimize authority arbitration.

\bibliographystyle{IEEEtran}
\bibliography{ref}

\end{document}